\documentclass[runningheads]{llncs}
\usepackage[T1]{fontenc}
\usepackage{graphicx}
\usepackage{amsmath}
\usepackage{amssymb}
\newcommand{\ra}[1]{\renewcommand{\arraystretch}{#1}}

\newcommand{\interval}[1]{{\scriptsize\textpm#1}}

\DeclareUnicodeCharacter{202C}{ }
\usepackage[pagebackref,breaklinks,colorlinks]{hyperref}
\usepackage{booktabs}
\usepackage[acronym,xindy,toc]{glossaries}
\usepackage{bm} 
\usepackage{tabularx} 
\usepackage{booktabs} 
\usepackage{makecell} 
\usepackage{mathtools}
\usepackage{multirow}

\begin{document}
\title{HALOS: Hallucination-free Organ Segmentation after Organ Resection Surgery}

\titlerunning{HALOS: Hallucination-free Organ Segmentation}
%
\author{Anne-Marie Rickmann\thanks{The authors contributed equally.} \inst{1,2} \and
Murong Xu\inst{\star 2} \and
Tom Nuno Wolf\inst{2} \and
Oksana Kovalenko\inst{2} \and Christian Wachinger\inst{1,2}}

%
\authorrunning{Rickmann et al.}
%
\institute{Lab for Artificial Intelligence in Medical Imaging, Ludwig Maximilians University, Munich, Germany \email{arickman@med.lmu.de} \and
Department of Radiology, Technical University Munich, Munich, Germany 
}

%
\maketitle              
\begin{abstract}
The wide range of research in deep learning-based medical image segmentation pushed the boundaries in a multitude of applications. 
A clinically relevant problem that received less attention is the handling of scans with irregular anatomy, e.g., after organ resection.
State-of-the-art segmentation models often lead to \emph{organ hallucinations}, i.e.,  false-positive predictions of organs, which cannot be alleviated by oversampling or post-processing.
Motivated by the increasing need to develop robust deep learning models, we propose HALOS for abdominal organ segmentation in MR images that handles cases after organ resection surgery.
To this end, we combine missing organ classification and multi-organ segmentation tasks into a multi-task model, yielding a classification-assisted segmentation pipeline. 
The segmentation network learns to incorporate knowledge about organ existence via feature fusion modules.
Extensive experiments on a small labeled test set and large-scale UK Biobank data demonstrate the effectiveness of our approach in terms of higher segmentation Dice scores and near-to-zero false positive prediction rate.  
\end{abstract}
\section{Introduction}
\begin{figure}[htpb!]
    \centering
    \includegraphics[width=\textwidth]{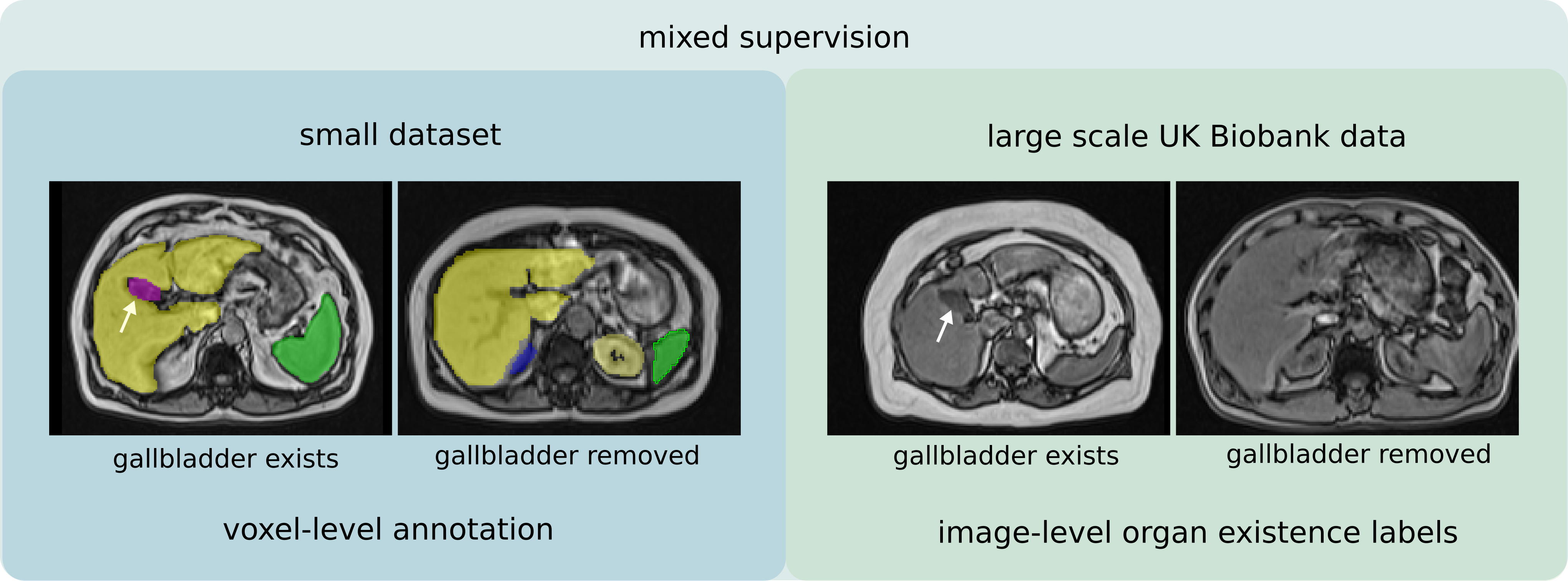}
    \caption{Mixed supervision in HALOS using a small dataset with voxel-level annotations of multiple organs and a large-scale dataset with image-level binary labels of organ existence. The white arrow points to the gallbladder.}\label{img: mixed_supervision}
    \end{figure}
Deep learning methods have become state-of-the-art for many medical image segmentation tasks, e.g. structural brain segmentation~\cite{roy2019}, tumor segmentation~\cite{liu2022deep} or abdominal organ segmentation~\cite{isensee2021nnu,roth2017hierarchical,chen2020fully,gibson2018automatic}.
A challenge that remains is the generalization to unseen data, where a domain shift between training and testing data often leads to performance degradation. 
Research on robustness and domain adaptation~\cite{guan2021domain} introduced new methods for handling domain shift, where the focus has mainly been on a shift in the intensity distribution of image data due to different imaging protocols, different scanner types or different modalities. 

In contrast, a domain shift in the anatomy itself, e.g., by missing organs due to surgical organ resection has received less attention. 
In comparison to natural images, which can show image compositions with arbitrary objects, medical images of the human abdomen usually contain the same organs in the same ordering. 
This constraint of the human anatomy is beneficial for training networks and has, for instance, been explicitly used by incorporating shape priors~\cite{zhou2019prior,oktay2017anatomically,liu2021anatomy}.
However, as we move to clinical translation or to large-scale population studies, we will also encounter cases that do not follow the normal anatomy, which will yield a degradation in segmentation accuracy. 

In this work, we mainly focus on gallbladder resection (cholecystectomy), as it is one of the most commonly performed abdominal surgeries. 
The indication for gallbladder removal is usually gallstones, which most of the time has no effect on other organs and the overall anatomy.
We further evaluate our method on the cases of nephrectomy (kidney resection), where the indication can be more severe, e.g., kidney tumors, which could come with anatomical changes in other organs, like metastases. 
Further, kidneys are much larger than gallbladders, so that their removal can lead to post-surgical organ shift~\cite{suzuki2012multi}.

As we will demonstrate in our experiments, state-of-the-art segmentation networks often identify organs in the images, although they were removed. 
A phenomenon that we refer to as \emph{organ hallucination}. 
We believe that organ hallucinations have so far not received more attention because publicly available segmentation datasets rarely contain cases after organ resection. 
This is probably due to the relatively small sample size of most segmentation datasets, as manual segmentation is time-consuming and costly.
Fortunately, large-scale population imaging studies like the UK Biobank (UKB) Imaging study~\cite{littlejohns2020uk} with a targeted 100,000 subjects are becoming available that provide representative data of the population. 
The prevalence of cholecystectomies (gallbladder resection) in our sample of UK biobank is 3.7$\%$, which provides enough data for studying this  research question. 
We introduce HALOS for the HALlucination-free Organ Segmentation after organ resection surgery. 
HALOS is a multi-task network that simultaneously learns classification of organ existence and segmentation of six abdominal organs (liver, spleen, kidneys, pancreas, gallbladder).
HALOS is trained using mixed supervision, which accounts for the fact that we only have voxel-level annotations for a small dataset but image-level labels of organ removal on a large dataset, see Figure~\ref{img: mixed_supervision}. 
A key component of HALOS is a feature fusion module~\cite{wolf2022daft} that integrates the knowledge of organ existence into the segmentation branch. 
The key contributions are:
\let\labelitemi\labelitemii
\begin{itemize}
    \item a robust and flexible multi-task segmentation and classification model that predicts near-to-zero false positive cases on the UKB dataset
    \item the multi-scale feature fusion with the dynamic affine feature map transform~\cite{wolf2022daft} of the classification output into the segmentation branch
    \item a demonstration of the relevance of the missing organ problem by comparing to state-of-the-art segmentation models. 
    
\end{itemize}

\subsection{Related Work}
\noindent
\textbf{Abdominal multi-organ segmentation}
Nowadays, convolutional neural networks are state-of-the-art for abdominal organ segmentation  in CT and MRI scans~\cite{roth2017hierarchical,chen2020fully,gibson2018automatic,wang2019abdominal,bobo2018fully}.
One method to point out is nnU-Net~\cite{isensee2021nnu}, which is an automatic pipeline to configure a U-Net to a given dataset. nnU-Net has won several medical image segmentation challenges, and has proven to be a robust and generic method. Therefore we consider nnU-Net as a baseline in our experiments.

\noindent
\textbf{Missing organ segmentation}
To our best knowledge, the missing organ problem has so far only been studied for CT scans in~\cite{suzuki2012multi}, where an atlas-based approach is used. 
It trains a Gaussian Mixture Model on normal images and detects missing organs by analyzing fitting errors. However, this method inevitably relies on heavy simulation for parameter tuning and is therefore vulnerable to distribution shift.
In a more recent method~\cite{tilborghs2022dice}, the Dice loss was studied and it was argued that setting the reduction dimension over the complete batch would help to predict images with missing organs. However, the method was not tested on cases after organ resection.
We compare to this approach in our experiments.

\noindent
\textbf{Classification-assisted segmentation}
As image-level labels are easier to obtain than voxel-wise annotations, prior work has considered including these additional labels by extending the segmentation network with a classification branch~\cite{wu2021jcs,mehta2018net,mlynarski2019deep}.
In~\cite{mlynarski2019deep}, the two branches are trained jointly using both fully-annotated and weakly-annotated data with shared layers at the beginning, for 2D brain tumor segmentation and classification of tumor existence. 
They showed that the additional classification significantly improved segmentation performance compared to standard supervised learning.
We compare to this approach in our experiments.

\noindent
\textbf{Feature Fusion}
Some approaches for classification-assisted segmentation use feature fusion, i.e., the interweaving of segmentation and classification branches. 
For example, separate segmentation and classification models are trained in~\cite{wu2021jcs} for Covid-19 diagnosis.
Feature maps of the classification and segmentation model are merged with Squeeze-and-Excitation (SE) blocks~\cite{hu2018squeeze}.
After the feature fusion, the enhanced feature map is fed into the decoder for segmentation.
An alternative for feature fusion is the combination with metadata, such as age, gender, or measurements of biomarkers. 
The Dynamic Affine Feature Map Transform (DAFT)~\cite{wolf2022daft} predicts the scales and shifts to excite or repress feature maps on a channel-level from such metadata, as seen in Fig.~\ref{img:pipeline}.

\section{Methods}
\begin{figure}[htpb!]
    \centering
    \includegraphics[width=\textwidth]{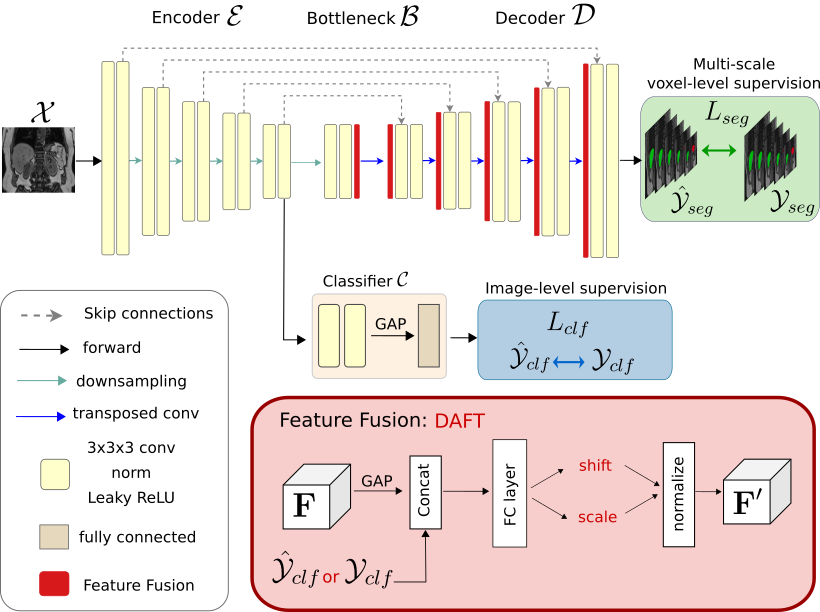}
    \caption{Overview of the HALOS multi-task pipeline.}\label{img:pipeline}
    \end{figure}

Figure \ref{img:pipeline} illustrates the dual-branch classification-assisted segmentation pipeline of HALOS that combines Multitask Learning and Feature Fusion to handle missing organs. 
In the following, we describe each part of our pipeline in detail.

\noindent
\textbf{Segmentation branch}
\label{sec:segmentation_branch}
In the segmentation branch, we use a U-Net architecture, based on nnU-Net~\cite{isensee2020} as the segmentation network which consists of an encoder $\mathcal{E}$, bottleneck $\mathcal{B}$, and a decoder $\mathcal{D}$.
As previously mentioned, nnU-Net~\cite{isensee2020} is one of the most generic and well-performing medical image segmentation models. 
The nnU-Net pipeline automatically determines the best U-Net architecture and data augmentation for the given data. 
Therefore, we fed our segmentation dataset into the nnU-Net pipeline and took the architecture of the best-performing  nnU-Net model and the data augmentation scheme as our baseline. The resulting model is a 3D U-Net with 32 starting channels and 5 downsampling levels. 

The advantage of using encoder-decoder structured networks is that intermediate representations can be obtained at different scales. 
The U-Net is trained on input MR images $\mathcal{X}$ and voxel-level annotations $\mathcal{Y}_{seg}$ to output segmentation predictions $\mathcal{\hat{Y}}_{seg}$ under full supervision. 
The segmentation loss is defined as an average of Dice and Cross-Entropy loss $L_{seg}$ with enabling of deep supervision at each feature map scale and dynamic class weights for individual images:
\begin{equation}
    \begin{aligned}
    &L_{seg} = L_{CE} + L_{Dice}, \quad \quad
    L_{CE}=-\frac{1}{N}\sum_i^Cy_i\log(\hat{y}_i),\\
    &L_{Dice} = 1-\frac{2\cdot|\hat{\mathcal{Y}}_{seg} \cap \mathcal{Y}_{seg}\vert + \epsilon}{|\hat{\mathcal{Y}}_{seg} \vert + |\mathcal{Y}_{seg} \vert + \epsilon }, 
    \label{eq: segloss}
\end{aligned}
\end{equation}

where we denote the class-wise ground truth $y_i$, class-wise predictions $\hat{y}_i$, the number of classes $C$ and samples $N$, a smoothing term $\epsilon$.
Note that some implementations of the Dice loss only add $\epsilon$ to the denominator, to avoid division by zero. In our case, it is important to add $\epsilon$ to numerator and denominator, as we want to ensure a Dice loss of $0$, rather than $1$, for true negative predictions of gallbladders.\\\\

\textbf{Classification branch}
Compared to manual voxel-level annotations, the global image-level labels are less informative but can be obtained at a substantially lower cost. 
Hence, we incorporate the classification task into the pipeline to study the impact of the low-dimensional prior knowledge on the final predicted segmentation.
In the classification branch, classifier $\mathcal{C}$ is built on top of the encoder $\mathcal{E}$ and takes a feature map from a specific encoder block as input.
The precise location of the classifier can be tuned as a hyper-parameter, but we found encoder blocks 4 and 5 promising for most models. 
Compared to training a standalone classification model, such a shared feature structure between $\mathcal{C}$ and $\mathcal{E}$ enables a more lightweight classification model and thus saves redundant computation. 
$\mathcal{C}$ consists of a convolutional block with the same structure as an encoder block, a 3D global average pooling step, and a fully connected layer for producing the final classification. 
The classifier is trained on MR scans with image-level surgery labels $\mathcal{Y}_{clf}$. 
The classification loss $L_{clf}$ is the average cross-entropy weighted by the actual class ratio in the training set.\\\\

\textbf{Feature fusion}
A key component of HALOS is the feature fusion module. 
The prior information about the resection of the gallbladder is fused with the feature maps of the segmentation branch at multiple locations. 
As shown in Figure \ref{img:pipeline}, these locations are the bottleneck and each stage of the decoder. 
Importantly, we can either use the ground truth image-level labels $\mathcal{Y}_{clf}$ or the classifier's prediction $\mathcal{\hat{Y}}_{clf}$ as input to the feature fusion, depending on whether the information about previous surgeries is available at test time.

We use DAFT~\cite{wolf2022daft} to perform feature fusion, which was originally designed to combine 3D images with low-dimensional tabular information, and can be conveniently integrated into any type of CNN. 
In our case, the tabular data to be concatenated is the binary classification result or ground truth label about gallbladder resection. 
To the best of our knowledge, DAFT has not yet been used in segmentation models or in a multi-scale fashion.
We expect that information sharing at multiple scales of the decoder will emphasize the prior knowledge about the organ’s presence and conduce the decoder to produce fewer false positive predictions of non-existing classes. 
The exact position of integrating feature fusion modules into the U-Net architecture is illustrated in Figure \ref{img:pipeline}. 
The classification labels are fused to the bottleneck feature map, which contains the highest-level information. 
Then the fused version will be forwarded to the decoder where we repeat the feature fusion blocks after each transpose convolution. 
We place feature fusion via DAFT before each decoder block, which avoids interaction with other normalization layers.
Formally, for each item in a batch, let $\mathbf{\hat{y}} \in \mathbb{R}$, be the predicted output from the classifier, and 
$\mathbf{F}_{d,c} \in \mathbb{R}^{D \times H \times W}$, where $D, H, W$ denote the depth, height, and width of the feature map, the $c$-th channel of the input feature map of block $d \in \{0,\ldots,5\}$ in the decoder, as illustrated in Figure~\ref{img:pipeline}.
DAFT~\cite{wolf2022daft} learns to predict scale $\alpha_{d,c}$ and offset $\beta_{d,c}$
\begin{align}
  \mathbf{F}^\prime_{d,c} &= \alpha_{d,c} \mathbf{F}_{d,c} + \beta_{d,c}, \\
  \alpha_{d,c} = f_c(\mathbf{F}_{d,c}, \mathbf{\hat{y}}_d),&\qquad
  \beta_{d,c}  = g_c(\mathbf{F}_{d,c}, \mathbf{\hat{y}}_d),
\end{align}
where $f_c$, $g_c$ are arbitrary mappings from image and tabular space to a scalar.
As proposed in~\cite{wolf2022daft}, a single fully connected neural network $h_c$ models $f_c$, $g_c$ and outputs a single $\alpha$-$\beta$-pair.

During training, we randomly sample MR images with voxel-level and image-level labels to form batches and use them to update the segmentation model and classifier respectively. 
With the previously defined $L_{seg}$ and $L_{clf}$, the final loss of HALOS is
\begin{equation} \label{eq: weightedloss}
    L = \alpha \cdot L_{seg} + (1 - \alpha) \cdot L_{clf},
\end{equation}
where $\alpha$ indicates the weight assigned to the segmentation loss.

\section{Results and Discussion}
\subsection{Experiment Setup}
\noindent
\textbf{Segmentation data}
We use whole-body MRI scans with voxel-level annotations from three different sources: the German National Cohort (NAKO)~\cite{bamberg2015whole}, the Cooperative Health Research in the Region of Augsburg (KORA)~\cite{bamberg2017subclinical}, and UKB~\cite{littlejohns2020uk}.
The samples cover a general population from Germany and the UK.
All three studies acquired abdominal images with a two-point Dixon sequence, where we use the oppose-phase scans in this work.
For pre-processing, we follow guidelines of other work~\cite{kart2022automated,rickmann2022abdomennet}.
The scans were manually segmented by a medical expert.
The dataset contains 63 scans in total (16 NAKO, 15 KORA, 32 UKB), of which 18 are patients after gallbladder resection. We have split this data into 42(9) scans for training, 7(3) for validation and 11(6) scans for testing, the count of missing gallbladder cases is given in parentheses. 

\noindent
\textbf{UKB data}
The UK Biobank dataset is much larger than the segmentation data, but only contains image-level annotations indicating organ presence.
We use it for training the organ existence classifier in our multi-task pipeline.
It can also be used for evaluating the model robustness since we can count the false positive segmentations of non-existing gallbladders.
We used the information about past surgeries from the UKB database and our medical expert verified the labels for correctness.
Out of 19,000 images we requested from UK Biobank, we counted 701 after gallbladder removal. We additionally randomly selected normal subjects.
We split the data into two subsets, one for training and validation of models (899 scans with and 349 without gallbladder), and one which serves as an unseen test set (952 scans with and 352 without gallbladder).
The ratio of no-gallbladder cases in each subset is set to be roughly 0.4.

\noindent
\textbf{Implementation details and hyperparameter tuning}
In this work, we use GPUs DGX A100 for running our experiments. The implementation is based on Python, PyTorch and MONAI. We perform hyperparameter tuning for the loss weight $\alpha$, weight decay, learning rates for the segmentation model and classifier, normalization type (instance or batch normalization), batch size, and the location of the classifier using Ray Tune. 
We train our models using the automated mixed precision of PyTorch.
Our code is publicly available at \url{https://github.com/ai-med/HALOS}.

\noindent
\textbf{Metrics}
We evaluate our models by comparing Dice scores for all organs and false positive rate (FPR) of gallbladder segmentations.
We define a sample as false positive if one or more voxels have been segmented as non-existing gallbladder.
As the Dice score is not defined for non-existing organs, we define it to be $1$ for true negative cases and $0$ for false positive cases. Therefore, we can observe large changes in the Dice score when reducing the false positive rate.

\noindent
\textbf{Baselines}
Apart from the nnU-Net baseline as described in Section~\ref{sec:segmentation_branch}, we further choose two alternative baselines, i.e., oversampling and post-processing.
For oversampling, we oversample the cases without a gallbladder in training to achieve a balance in class frequency. 
Note that we are already weighting the loss functions by class frequency.
The post-processing baseline is another method, where we use the prior information about gallbladder resection to remove false positives as a direct post-processing step. 

\begin{table*}[t]
\footnotesize
\setlength{\tabcolsep}{2pt} 
\caption{Comparison of HALOS with baseline nnU-Net, with oversampling, post-processing, Dice loss with batch reduction\cite{tilborghs2022dice} (Dice batch red.), and multi-task model~\cite{mlynarski2019deep}.  
FF: feature fusion, gt: ground truth labels at test-time. We list Dice scores for all organs and false positive rate (FPR) for removed gallbladders.
We provide mean and standard deviation over 5-fold cross-validation.\\
*:  best architecture for our data proposed by the nnU-Net pipeline was re-implemented.}
\ra{1.3}
\resizebox{\linewidth}{!} {

\begin{tabular}{l c c c c c c c c}
    \toprule
    & \multicolumn{7}{c}{Dice Scores $\uparrow$} & \multirow{2}{*}{FPR $\downarrow$} \\
     \cmidrule(lr{1em}){2-8}
      \addlinespace[0.3em]

    Method & Mean & liver & spleen & r kidney & l kidney & pancreas & gallbl. & \\

    \midrule
    
    
    nnU-Net*\cite{isensee2020} & $0.823$\interval{0.014} & $0.938$\interval{0.004}& $0.891$\interval{0.006}& $0.898$\interval{0.003} & $0.894$\interval{0.002} & $0.643$\interval{0.016} & $0.674$\interval{0.076}& $0.267$\interval{0.149}\\
    \addlinespace[0.3em]

    + oversampling & $0.832$\interval{0.008}& $0.940$\interval{0.006} & $0.894$\interval{0.005}& $0.901$\interval{0.005} & $0.891$\interval{0.005} & $0.655$\interval{0.011}& $0.712$\interval{0.052} & $0.233$\interval{0.091}\\
  \addlinespace[0.3em]
  
    + post-proc. (gt) & $0.847$\interval{0.005} & $0.938$\interval{0.004} & $0.891$\interval{0.006} & $0.898$\interval{0.003} & $0.894$\interval{0.002} & $0.643$\interval{0.016} & $0.819$\interval{0.009} & $0$\interval{0}\\

  \addlinespace[0.3em]
  
    + batch red.~\cite{tilborghs2022dice} & $0.818$\interval{0.010} & $0.945$\interval{0.002} & $0.895$\interval{0.002} & $0.901$\interval{0.005}& $0.894$\interval{0.006} & $0.663$\interval{0.014}& $0.610$\interval{0.045}& $0.400$\interval{0.091} \\
  \addlinespace[0.3em] 
      \midrule

    \addlinespace[0.3em]

  multi-task~\cite{mlynarski2019deep} & $0.822$\interval{0.010} & $0.930$\interval{0.006} & $0.879$\interval{0.004} & $0.895$\interval{0.003} & $0.885$\interval{0.002} & $0.625$\interval{0.016} & $0.716$\interval{0.054} & $0.233$\interval{0.091} \\
  \addlinespace[0.3em]

    HALOS w/o FF & $0.825$\interval{0.010} & $0.941$\interval{0.002} & $0.892$\interval{0.009} & $0.898$\interval{0.004} & $0.892$\interval{0.005} & $0.657$\interval{0.013} & $0.668$\interval{0.073} & $0.3$\interval{0.139} \\
    \addlinespace[0.3em]

  HALOS (pred, gt) & $0.853$\interval{0.002}& $0.939$\interval{0.003} & $0.899$\interval{0.005} & $0.899$\interval{0.003} & $0.893$\interval{0.004}& $0.649$\interval{0.021}& $0.840$\interval{0.015}& $0$\interval{0}\\


  \addlinespace[0.3em]
    \bottomrule
\end{tabular}}
\vspace{-0.5em}
\label{tab:comparison_seg}
\end{table*}

\begin{table*}[t]
\footnotesize
\setlength{\tabcolsep}{4.6pt} 
\caption{Comparison of HALOS with baseline nnU-Net, oversampling, post-processing, Dice loss with batch reduction~\cite{tilborghs2022dice} (+ batch red.) and multi-task model~\cite{mlynarski2019deep}  on the UKB dataset. FF: feature fusion, gt: ground truth labels for FF at test-time, pred:  classification predictions for FF at test-time. We provide false positive (FP), false negative (FN), true positive (TP), true negative (TN), false positive rate (FPR) and F1 score for removed gallbladders, and the balanced accuracy (BAcc) of all classifiers.
All values are mean and standard deviation over 5-fold cross-validation.\\
*:  best architecture for our data proposed by the nnU-Net pipeline was re-implemented.}
\ra{1.3}
\resizebox{\linewidth}{!} {
\begin{tabular}{l c c c c c c c}
    \toprule
    Method & FP $\downarrow$ & TN $\uparrow$ & TP $\uparrow$ & FN $\downarrow$ & FPR $\downarrow$ & F1 $\uparrow$& BAcc $\uparrow$\\
    \midrule
    
    
    nnU-Net*\cite{isensee2020} &  $91.2$ \interval{30.62} & $260.8$ \interval{30.62} & $537.2$ \interval{16.62} & $62.8$ \interval{16.62}  &$0.259$ \interval{0.087}& $0.875$ \interval{0.009} &  \\
    
    \addlinespace[0.3em]
    
    + oversampling & $66.6$ \interval{6.633} & $285.4$ \interval{6.633} & $522.6$ \interval{13.18} & $77.4$ \interval{13.18} & $0.189$ \interval{0.028} & $0.879$ \interval{0.011} &  \\

  \addlinespace[0.3em]
  
   + post-proc. (gt) & $ 0 $ \interval{0} & $352$ \interval{0} & $537.2$ \interval{16.62} & $62.8$ \interval{16.62}  &$0$ \interval{0}& $0.945$ \interval{0.015} & \\

  \addlinespace[0.3em]
  
    + batch red.~\cite{tilborghs2022dice} & $135.2$ \interval{57.15} & $216.8$ \interval{57.15} & $530.2$ \interval{24.39} & $69.8$ \interval{24.39} & $0.384$ \interval{0.162} & $0.838$ \interval{0.017} \\

    \midrule

  \addlinespace[0.3em] 
  
  multi-task~\cite{mlynarski2019deep} & $100.2$ \interval{16.48} & $251.8$ \interval{16.48} & $578.2$ \interval{3.701} & $21.8$ \interval{3.701} & $0.285$ \interval{0.047} & $0.905$ \interval{0.054} & $0.874$ \interval{0.045}  \\

  \addlinespace[0.3em]

 HALOS w/o FF & $52.6$ \interval{17.67} & $299.4$ \interval{17.67} & $547.4$ \interval{22.39}& $52.6$ \interval{22.39}& $0.149$ \interval{0.050} & $0.869$ \interval{0.056} & $0.896$ \interval{0.047} \\
  \addlinespace[0.3em]

  HALOS (gt) & $2$ \interval{2.550}& $350$ \interval{2.550} & $564.8$ \interval{14.74} & $35.2$  \interval{14.74} & $0.006$  \interval{0.007} & $0.968$  \interval{0.010} & $0.933$  \interval{0.005} \\
  \addlinespace[0.3em]

    HALOS (pred) & $11$   \interval{5.339}& $341$  \interval{5.339} & $541.8$  \interval{14.20}& $58.2$  \interval{14.20} & $0.031$  \interval{0.015}& $0.940$  \interval{0.010} &  $0.933$  \interval{0.005}\\

  \addlinespace[0.3em]
    \bottomrule
\end{tabular}}
\vspace{-0.5em}
\label{tab:comparison_ukb}
\end{table*}

\subsection{Experiments on cholecystectomy cases}
\begin{figure}[ht!]
    \centering
    \includegraphics[width=\textwidth]{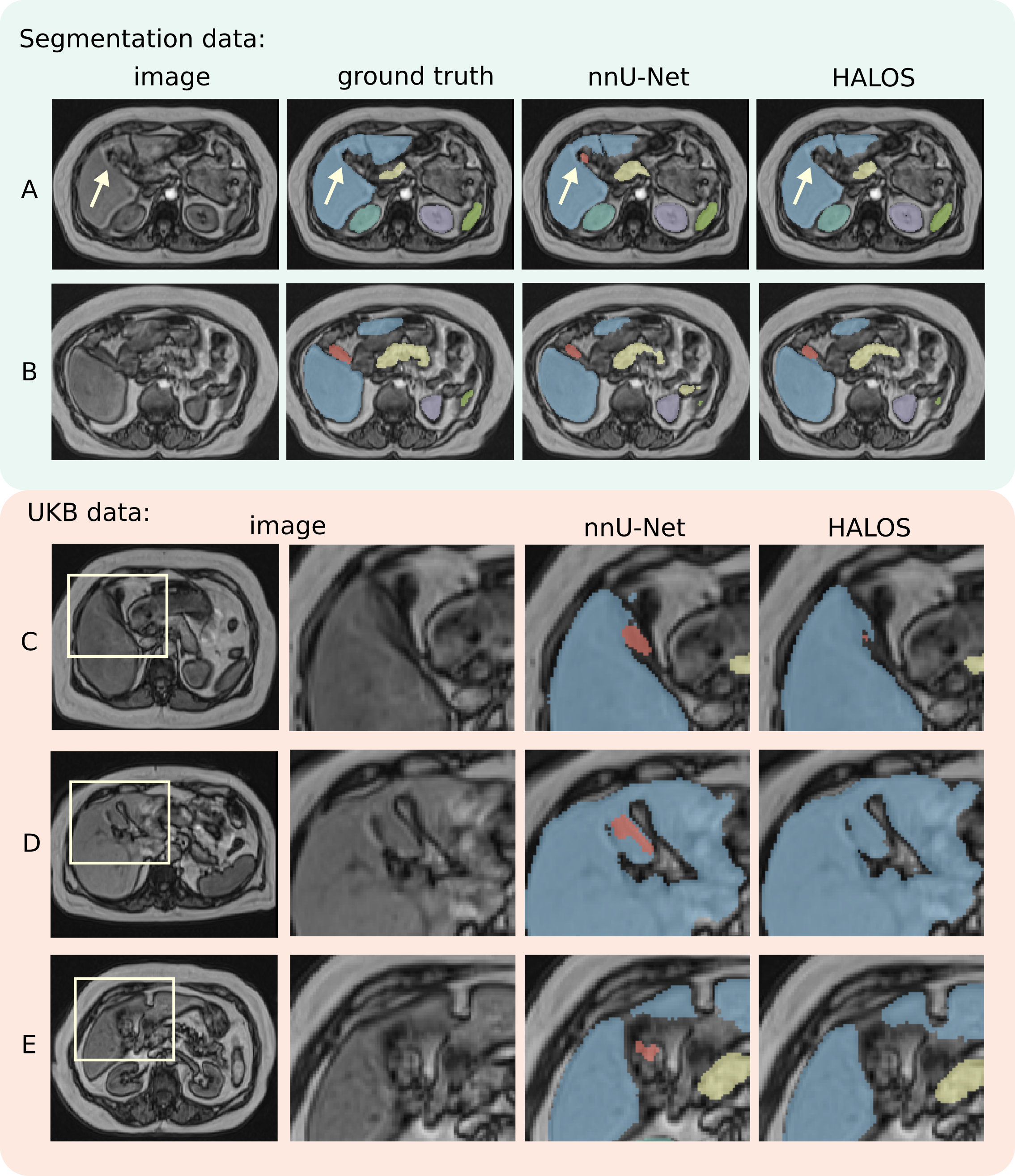}
    \caption{Segmentation results on the segmentation data (top) and UKB  (bottom). Comparison of nnU-Net and HALOS. 
    A: scan with a resected gallbladder, nnU-Net produces a false positice.
    B: scan with an existing gallbladder.
    C: both models predict a false positive in the location where the gallbladder was resected.
    D: nnU-Net produces a false positive inside the liver.
    E: nnU-Net produces a false positive in the intestine.}\label{fig:segmentation}
    \vspace{-0.5em}
    \end{figure}

We train the baseline nnU-Net, oversampling and post-processing baselines, state-of-the-art methods~\cite{tilborghs2022dice,mlynarski2019deep} and HALOS using 5-fold cross-validation and report the average results over all folds on the segmentation test set in table~\ref{tab:comparison_seg} and on the UKB test set in table~\ref{tab:comparison_ukb}.
The average FPR is quite high for the baseline nnU-Net on both datasets, which leads to a low gallbladder Dice score of $0.674$. The segmentation performance on pancreas is also low $0.643$, but the pancreas is very hard to segment, due to its shape variability.
The oversampling only slightly improves performance, so we can assume that the reason for the high FPR is not only caused by class imbalance.
As expected, the post-processing leads to higher gallbladder Dice scores and zero FPR, since it uses the ground truth information about cholecystectomy. 
A shortcoming of the post-processing is that the model’s false positive prediction may appear in neighboring organs, which will result in a hole in the segmentation. 
The gallbladder usually lies in fossa vesicae biliaris, which is a depression on the visceral surface of the liver anteriorly, between the quadrate and the right lobes.
Since the location is closely connected to the liver, we found many mistakes produced by our baseline that are either localized inside the liver or partly in the liver and partly in other tissues like visceral fat.
Examples of typical organ hallucinations are shown in Figure~\ref{fig:segmentation} C-D, where the gallbladder is predicted in the fossa vesicae biliaris (C), inside the liver (D) and in the intestine (E).
The recent work~\cite{tilborghs2022dice} proposes to set $\epsilon$ in the Dice loss to a low value, e.g. $10^{-7}$, the batch size higher than $1$ and to reduce the Dice loss over the batch dimension. We set the batch size to $8$, which reached the limit of our GPU memory. 
Note that in our baseline model the batch size is set to $2$, $\epsilon$ is $1$ and we also reduce over the batch dimension.
Interestingly, we observe an increase in FPR for both datasets. 
In preliminary experiments, we have removed the batch reduction in the Dice loss, but we have observed no significant difference in performance.
The multi-task model proposed in~\cite{mlynarski2019deep} includes a classifier right before the segmentation output of the decoder. We use our nnU-Net model and extend it with a classifier, following the architecture proposed in~\cite{mlynarski2019deep} at decoder block 5. 
This model leads to a slight decrease in FPR on the segmentation data, but to a higher FPR on the UKB data. 
To analyze the impact of multi-task learning and the feature fusion models, we train HALOS without the feature fusion, which interestingly leads to a slight increase in FPR and an decrease in gallbladder Dice score and slight decrease in FPR on UKB, compared to nnU-Net, even though the balanced accuracy of the classifier is $0.896$. 
Therefore we argue, that multi-task training alone is not sufficient to reduce organ hallucinations.
HALOS was trained using the ground truth labels $\mathcal{Y}_{clf}$ as input for feature fusion, and leads to an impressive reduction of the FPR to $0$ on the segmentation data and $0.006$ on the UKB data. 
The multi-task classifier achieves a balanced accuracy of $0.93$.
When we use the classifier's prediction for feature fusion at test time, we observe a slight increase in FPR over using the ground truth labels to $0.03$.
This shows, that our method is flexible and depending if prior information about gallbladder resection is available at test time or not, one can either fuse the ground truth labels or the classifier's predictions. 

\subsection{Experiments on nephrectomy cases}
To evaluate if HALOS can be applied to other organ resection cases, we validate the effectiveness of HALOS on cases after nephrectomy. 
Note that we did not do any further hyper-parameter tuning in this experiment.
We create a kidney segmentation dataset that contains 46(6/2) scans for training and 10(2/1) for testing, the count of missing kidney cases is given in parentheses with the format left/right. For UKB data, we split the available subjects into one training and validation set (200 scans with 17/5 missing left/right kidneys), and another hold-out test set (55 scans with 4/2 no-kidneys). 
Similar to gallbladder experiments, we train the baseline nnU-Net and HALOS using 5-fold cross-validation. 
Note that the classifier learns a multi-class classification, in contrast to the binary classification in the gallbladder experiments.
We report the results of the nephrectomy experiment in the following.
The baseline nnU-Net achieved an FPR of  $0.2$ for left kidney and $1$ for right kidney on the UKB data.
We observe that HALOS achieves a lower FPR of $0$ for the left kidney and still high FPR of $0.7$ for the right kidney, while having a significantly reduced voxel-level FP count of $16.5$, compared to $129$ for nnU-Net. 
The left kidney Dice of HALOS ($0.9024$) is higher than of nnU-Net ($0.8413$) while having no improvement on the right kidney $0.864$ vs $0.867$.
A possible reason might be the small dataset size with severe class imbalance, of having only two cases with missing right kidneys in the training set. 
The balanced accuracy of the HALOS classifier is $0.93$ for left kidney and $0.58$ for right kidney, which also suggests that the class imbalance has more impact in this setting.

\section{Conclusion}
In this work, we introduced HALOS, a multi-task classification and segmentation model for hallucination-free organ segmentation. 
We propose to use multi-scale feature fusion, via the dynamic affine feature-map transform, to enrich the feature maps of the segmentation branch with prior information on organ existence.
We have shown on cases after cholecystectom0,y and nephrectomy, that HALOS significantly reduces false positive predictions on a large scale UK Biobank test set, and increases gallbladder and left kidney Dice scores on a smaller segmentation test set, compared to nnU-Net and several additional baselines and multi-task approaches.
HALOS is flexible to use ground truth organ existence labels at test-time or the prediction of the classifier, depending on the availability of such labels.
In future work we would like to extend HALOS to additional cases of organ resection, e.g. hysterectomy (removal of uterus) or splenectomy (removal of spleen).

\paragraph{Acknowledgment}
This research was partially supported by the Bavarian State Ministry of Science and the Arts and coordinated by the bidt, the BMBF  (DeepMentia, 031L0200A), the DFG and the LRZ.

\bibliographystyle{splncs04}
\bibliography{references}
\end{document}